\documentclass[letterpaper, 10 pt, conference]{ieeeconf}  % Comment this line out if you need a4paper
\IEEEoverridecommandlockouts                              % This command is only needed if 
\usepackage[noadjust]{cite}
\usepackage{color}
\usepackage{lipsum}
\usepackage{siunitx}
\usepackage{booktabs}
\usepackage{multirow} 
\usepackage{graphicx}
\usepackage{caption}
\usepackage{tabularx}
\usepackage{threeparttable}
\usepackage{threeparttablex}
\usepackage{bbm}
\usepackage{float}
\let\labelindent\relax
\usepackage{enumitem} 
\usepackage{amsmath}
\usepackage{amssymb}
\usepackage{arydshln}

\usepackage[linesnumbered,ruled,vlined]{algorithm2e}

\usepackage{changepage}
\usepackage{bbm}
\usepackage{siunitx}
\usepackage{bm}
\usepackage{color}
\usepackage{graphbox,graphicx}
\usepackage{stfloats}
\usepackage{amsmath}
\usepackage{amssymb}
\usepackage{calligra}
\usepackage{bbm}
\usepackage{array}
\usepackage{multirow}
\usepackage{tabularray}
\usepackage{makecell}
\usepackage{hhline}
\usepackage[utf8]{inputenc}
\usepackage[caption=false,font=normalsize,labelfont=sf,textfont=sf]{subfig}
\usepackage{fixltx2e}
\usepackage{kotex}
\usepackage{float}
\usepackage{eufrak}
\usepackage{supertabular}
\usepackage[table]{xcolor} 
\usepackage{tabularx}
\usepackage{soul}
\usepackage{enumitem}
\usepackage{svg}
\usepackage{url}
\usepackage{booktabs}
\usepackage{threeparttable}
\usepackage{pifont}
\usepackage{dsfont}
\usepackage[colorlinks=true, urlcolor=blue, linkcolor=red]{hyperref}
\usepackage{gensymb}
\usepackage{cancel}

\title{\LARGE \bf
Uncertainty-Aware Non-Prehensile Manipulation \\
with Mobile Manipulators under Object-Induced Occlusion
}

\author{
    Jiwoo Hwang$^{1}$, Taegeun Yang$^{2}$, Jeil Jeong$^{1}$, Minsung Yoon$^{2}$ and Sung-Eui Yoon$^{2\dagger}$%
    \thanks{$^{1}$J. Hwang and J. Jeong are with the Robotics Program at the Korea Advanced Institute of Science and Technology (KAIST), Daejeon, 34141, Republic of Korea.}%
    \thanks{$^{2}$T. Yang, M. Yoon and S. Yoon are with the School of Computing  at the same institute, KAIST.}%
    \thanks{$^{\dagger}$S.Yoon is a corresponding author. {\tt\small sungeui@kaist.edu}}%
}

\begin{document}
%%%%%%%%%%%%%%%%%%%%%%%%%%%%%%%%%%%%%%%%%%%%%%%%%%%%%%%%%%%%%%%%%%%%%%%%%%%%%%%%
\maketitle
\thispagestyle{empty}
\pagestyle{empty}
%%%%%%%%%%%%%%%%%%%%%%%%%%%%%%%%%%%%%%%%%%%%%%%%%%%%%%%%%%%%%%%%%%%%%%%%%%%%%%%%
\begin{abstract}

Non-prehensile manipulation using onboard sensing presents a fundamental challenge: the manipulated object occludes the sensor's field of view, creating occluded regions that can lead to collisions.
We propose \emph{CURA-PPO}, a reinforcement learning framework that addresses this challenge by explicitly modeling uncertainty under partial observability.
By predicting collision possibility as a distribution, we extract both risk and uncertainty to guide the robot's actions.
The uncertainty term encourages active perception, enabling simultaneous manipulation and information gathering to resolve occlusions.
When combined with confidence maps that capture observation reliability, our approach enables safe navigation despite severe sensor occlusion.
Extensive experiments across varying object sizes and obstacle configurations demonstrate that \emph{CURA-PPO} achieves up to $3\times$ higher success rates than the baselines, with learned behaviors that handle occlusions.
Our method provides a practical solution for autonomous manipulation in cluttered environments using only onboard sensing.

\end{abstract}

%%%%%%%%%%%%%%%%%%%%%%%%%%%%%%%%%%%%%%%%%%%%%%%%%%%%%%%%%%%%%%%%%%%%%%%%%%%%%%%%% 
\section{INTRODUCTION}
Non-prehensile manipulation, which controls objects without grasping, is commonly used to handle items that exceed grasping capabilities due to size, weight, or surface properties~\cite{ruggiero2018nonprehensile, dengler2022learning, dengler2024learning, ferrandis2023nonprehensile, jeon2023learning, feng2024learning, tang2023unwieldy, ozdamar2024pushing, dadiotis2025dynamic}.
Mobile platforms—including quadrupeds~\cite{jeon2023learning}, wheeled robots~\cite{tang2023unwieldy}, and mobile manipulators~\cite{dadiotis2025dynamic}—extend these capabilities to larger workspaces, enabling tasks such as transporting boxes in warehouses or repositioning furniture.
However, a fundamental challenge arises when these platforms rely solely on onboard sensors (e.g., LiDAR): the manipulated object itself occludes the sensors' fields of view, creating dangerous blind regions during manipulation.
This occlusion-induced uncertainty becomes critical when unexpected obstacles appear—such as items falling from shelves or personnel entering the workspace—thereby requiring real-time perception and avoidance.
Object-induced occlusion in non-prehensile tasks remains underexplored in the literature, despite its practical importance for safe autonomous manipulation in cluttered environments with only local sensing.

To address the perceptual challenges posed by object-induced occlusion, we leverage uncertainty-aware decision-making strategies.
Recent work has shown that when agents explicitly quantify uncertainty and incorporate its reduction as an objective, they exhibit information-seeking behaviors—actively exploring ambiguous regions rather than relying on incomplete observations~\cite{pan2022activenerf, georgakis2022uncertainty, jin2023neu, dengler2025efficient, jin2025learning}.
This principle directly applies to our setting: by incorporating\;uncertainty reduction alongside the manipulation objective, the robot proactively resolves occlusions while maintaining task progress.
Such uncertainty-driven exploration proves essential when the manipulated object blocks sensor views, as it encourages the robot to strategically maneuver for better observability without neglecting its manipulation goal.

Building on these principles, we propose \emph{CURA-PPO} (Collision Uncertainty-Risk Aware PPO), a reinforcement learning (RL) framework that enables safe non-prehensile manipulation with local sensing under occlusions.

\begin{figure}[t]
    \vspace{0.2cm}
  \centering
  \includegraphics[width=0.49\textwidth]{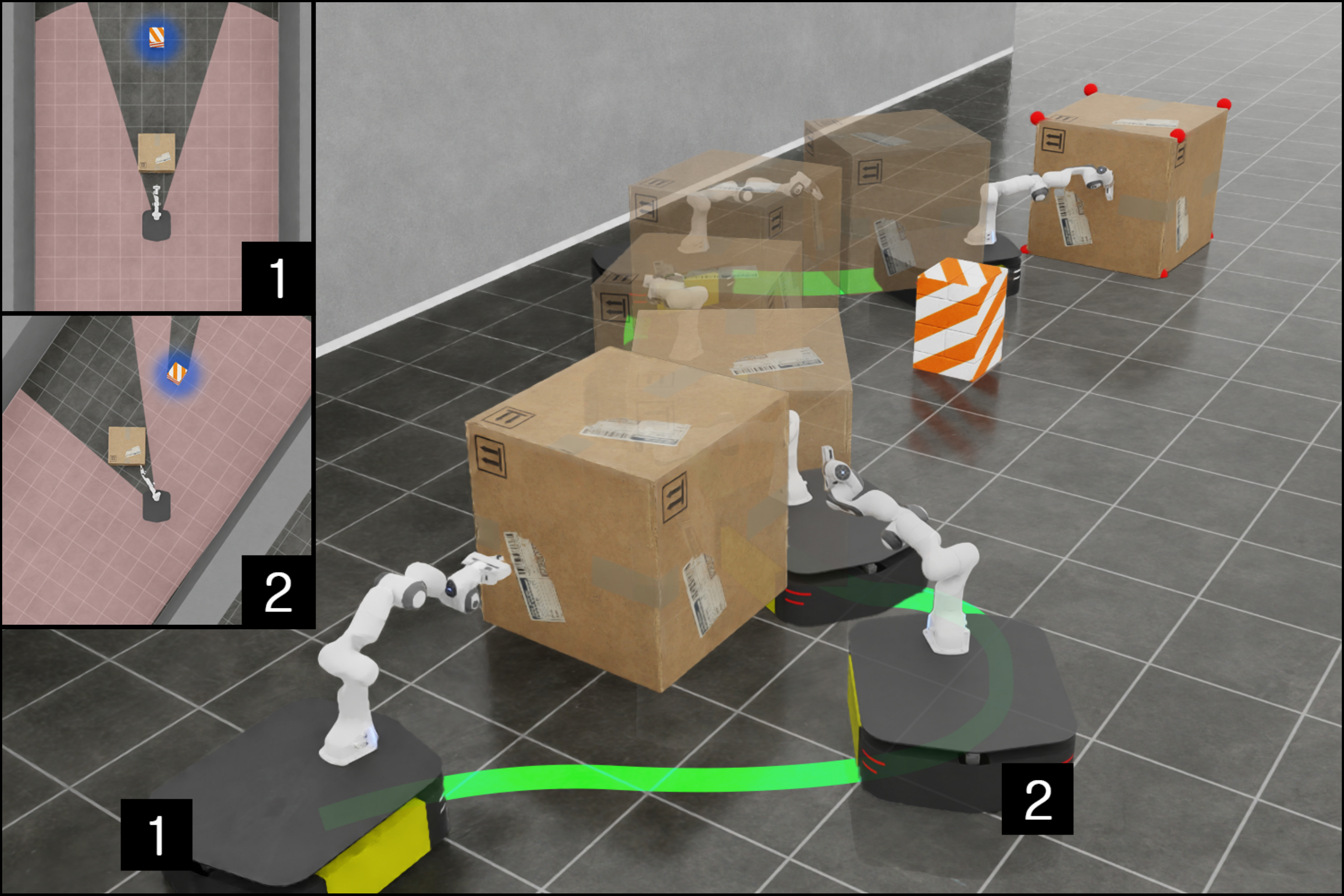}
    \caption{\small \textbf{Uncertainty-Aware Non-prehensile Manipulation under Object-Induced Occlusion}. A mobile manipulator pushes an object to the target using local onboard sensing. (1) The object induces occlusion, hiding an unforeseen obstacle. (2) The mobile manipulator adjusts its behavior to actively reduce uncertainty from occlusion, revealing the hidden obstacle for safe non-prehensile manipulation.}
  \label{fig:main}
  \vspace{-0.7cm}
\end{figure}

\noindent In summary, our main contributions are \emph{threefold}:
\begin{itemize}[leftmargin=0.4cm]
    \item We formulate non-prehensile manipulation under object-induced occlusion as an uncertainty-aware decision-making problem, where a mobile manipulator manages cluttered environments using only an onboard 2D LiDAR while the manipulated object creates occluded regions.
    \item We propose \emph{CURA-PPO}, an RL framework incorporating a Distributional Collision Estimator (\emph{DCE}) that predicts collision possibility as a distribution. By integrating the distribution's mean (\emph{risk}) and variance (\emph{uncertainty}) as intrinsic costs, our approach jointly optimizes task completion, collision avoidance, and uncertainty reduction.
    \item We demonstrate through extensive simulation experiments that \emph{CURA-PPO} significantly outperforms the baselines across varying object sizes and obstacle configurations, with qualitative analysis revealing learned active perception behaviors that proactively resolve occlusions.
\end{itemize}

%%%%%%%%%%%%%%%%%%%%%%%%%%%%%%%%%%%%%%%%%%%%%%%%%%%%%%%%%%%%%%%%%%%%%%%%%%%%%%%%
\begingroup
\linespread{0.99}\selectfont
\setlength{\parskip}{0pt}
\section{RELATED WORK}
\label{sec:2}

\subsection{Non-Prehensile Manipulation}
\label{sec:2-A}
\vspace{-0.25\baselineskip}
\emph{Non-prehensile} manipulation refers to a class of interactions that control an object by applying unilateral constraints (e.g., pushing), without the force closure provided by a grasp's bilateral constraints~\cite{ruggiero2018nonprehensile}.
This approach is effective for manipulating objects beyond a robot's physical limits, such as its payload capacity or grasp range~\cite{dengler2022learning, dengler2024learning, ferrandis2023nonprehensile, jeon2023learning, feng2024learning, tang2023unwieldy, ozdamar2024pushing, dadiotis2025dynamic}. 
Building on these advantages, research in non-prehensile manipulation spans tasks at varying scales, from precise object handling on tabletops with fixed-base arms~\cite{dengler2022learning, dengler2024learning, ferrandis2023nonprehensile} to long-range manipulation in extended environments with mobile robots~\cite{jeon2023learning, feng2024learning, tang2023unwieldy, ozdamar2024pushing, dadiotis2025dynamic}.

For long-range manipulation, prior work has shown promising results, yet it assumes complete knowledge of all obstacles in the environment, including their locations and geometries.
In practical scenarios where such information is not readily available, robots rely on onboard sensors.
This setting introduces challenges from occlusions, as the manipulated object blocks the sensors' views and limits environmental awareness.
To address these challenges, we propose an uncertainty-aware manipulation policy under partial observability from onboard sensors.
Specifically, our approach incorporates active sensing---purposeful movements to gather information---to handle uncertainty from occlusions while balancing uncertainty reduction with task execution.

\subsection{Uncertainty-Aware Decision-Making}
\label{sec:2-B}
\vspace{-0.25\baselineskip}
Decision making in partially observable environments requires handling uncertainty from incomplete sensing~\cite{kochenderfer2015decision}.
Such uncertainty has been quantified as the variance of model predictions, either through distributional representation learning~\cite{bellemare2017distributional, dabney2018distributional, amini2020deep}, or sampling-based approximations~\cite{gal2016dropout, lakshminarayanan2017simple, an2021uncertainty}.\newline
To enhance robustness under partial observability, recent work incorporates uncertainty estimates into decision making~\cite{rosman2016bayesian, da2020uncertainty, roemer2023vision, dass2024learning}.
Moreover, using uncertainty reduction as a learning signal encourages information-seeking behaviors, enabling agents to recognize and resolve perceptual ambiguities~\cite{pan2022activenerf, georgakis2022uncertainty, jin2023neu, dengler2025efficient, jin2025learning}.
We apply this principle to non-prehensile manipulation with local sensing, where occlusions reduce observability.
Our approach quantifies uncertainty through variance in collision possibility prediction: pushing toward observable regions yields low variance, while pushing into occluded areas produces high variance.
Incorporating this variance into policy objective, we encourage active perception behaviors that reduce occlusions during manipulation.
\endgroup
%%%%%%%%%%%%%%%%%%%%%%%%%%%%%%%%%%%%%%%%%%%%%%%%%%%%%%%%%%%%%%%%%%%%%%%%%%%%%%%%
\section{Variable Notations}
\label{sec:variable}
This section summarizes the key notations and variables used throughout this paper.
Cartesian position, linear velocity, and linear acceleration are denoted as $\boldsymbol{p}$, $\boldsymbol{v}$, and $\dot{\boldsymbol{v}} \in \mathbb{R}^3$.
XYZ Euler angles $\boldsymbol{\theta}$ represent the orientation, with angular velocity $\boldsymbol{\omega} \in \mathbb{R}^3$.
For the seven-DOF manipulator, joint position, velocity, and acceleration are $\boldsymbol{q}$, $\dot{\boldsymbol{q}}$, and $\ddot{\boldsymbol{q}} \in \mathbb{R}^{7}$.
Superscripts denote reference frames, while subscripts indicate entities, axis components, and time steps.
For example, ${p}^{b}_{g,x, t}$ represents the $x$-component of the goal position in the base frame at time step $t$, where $w$, $b$, $e$, $o$, and $g$ denote the world, base, end-effector, object, and goal frames, respectively.

%%%%%%%%%%%%%%%%%%%%%%%%%%%%%%%%%%%%%%%%%%%%%%%%%%%%%%%%%%%%%%%%%%%%%%%%%%%%%%%%
\section{PRELIMINARIES}
\label{sec:preliminaries}
This section provides the necessary background on Partially Observable Markov Decision Process (POMDP) and Proximal Policy Optimization (PPO) for understanding the subsequent methodological contributions.
In contrast to fully observable MDPs, a POMDP is defined by the tuple $(\mathcal{S}, \mathcal{A}, \mathcal{T}, \mathcal{R}, \Omega, \mathcal{O}, \gamma)$, where agents receive observations $\boldsymbol{o} \in \Omega$ rather than direct access to states $\boldsymbol{s} \in \mathcal{S}$.
Here, $\mathcal{A}$ is the action space, $\mathcal{T}$ the transition dynamics, $\mathcal{R}$ the reward function, $\mathcal{O}$ the observation function mapping states into the observation space $\Omega$, and $\gamma$ the discount factor.

The goal of reinforcement learning is to learn a policy $\pi_\phi(\boldsymbol{a}_t|\boldsymbol{o}_t)$ that maximizes the expected cumulative reward.
PPO~\cite{schulman2017proximal} optimizes this objective through iterative policy improvement using a surrogate objective:
\begin{equation}
\mathcal{J}^{PPO}(\phi) = \mathbb{E}_{(\boldsymbol{o}_t,\boldsymbol{a}_t) \sim \pi_{\phi_{\text{old}}}}\left[\frac{\pi_\phi(\boldsymbol{a}_t|\boldsymbol{o}_t)}{\pi_{\phi_{\text{old}}}(\boldsymbol{a}_t|\boldsymbol{o}_t)} \cdot \hat{A}_t\right]
\label{eq:ppo_standard}
\end{equation}
where the probability ratio $r(\phi) = \frac{\pi_\phi(\boldsymbol{a}_t|\boldsymbol{o}_t)}{\pi_{\phi_{\text{old}}}(\boldsymbol{a}_t|\boldsymbol{o}_t)}$ measures policy change.
In practice, the probability ratio is clipped to encourage stable policy updates.
The advantage function $A(\boldsymbol{s}_t, \boldsymbol{a}_t) = Q(\boldsymbol{s}_t, \boldsymbol{a}_t) - V(\boldsymbol{s}_t)$ compares the value of taking a specific action to the average value at that state, where $Q$ and $V$ represent action-value and state-value functions, respectively.
During training, the estimated advantage $\hat{A}_t$ serves as a learning signal in policy optimization~\cite{schulman2015high}: positive values encourage the policy to increase the probability of the chosen action, whereas negative values decrease it.

\section{METHOD}
\label{sec:method}
\begin{figure*}[t]
  \centering
  \includegraphics[width=2\columnwidth]{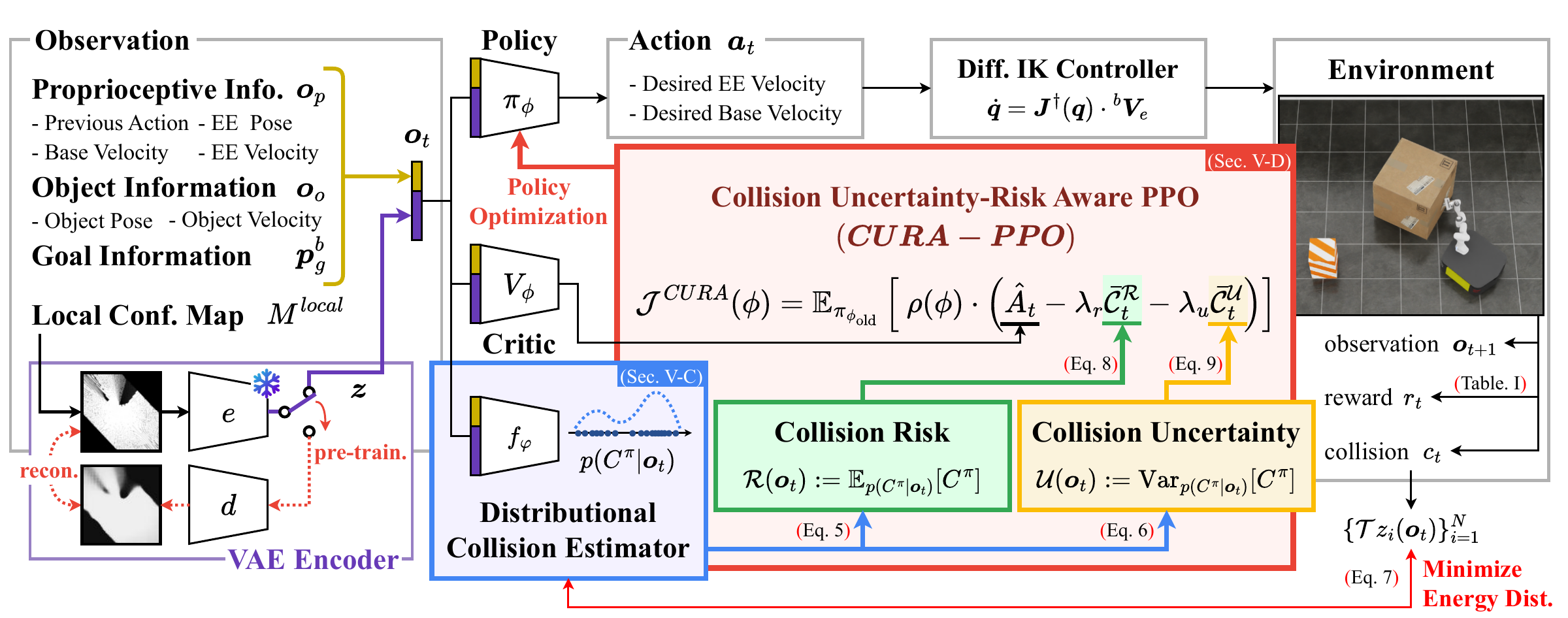}
  \caption{
      \small \textbf{Overall Framework.} Local observations including proprioceptive, object, goal, and confidence map features are processed by the policy network $\pi_{\phi}$ (see Sec.~\ref{sec:push}), which outputs velocity commands executed through differential inverse kinematics.
      The Distributional Collision Estimator (DCE, $f_{\varphi}$) predicts collision possibility as a distribution, extracting risk and uncertainty signals (see Sec.~\ref{sec:dce}) that augment the PPO objective in \emph{CURA-PPO}, encouraging behaviors that reduce both collision risk and perceptual uncertainty (see Sec.~\ref{sec:cura}).
  }
  \label{fig:framework}
  \vspace{-0.5cm}
\end{figure*}
We present \emph{Collision Uncertainty-Risk Aware PPO (CURA-PPO)}, a modified reinforcement learning algorithm for long-range non-prehensile manipulation under occlusions caused by the manipulated object itself.
Our approach is composed of two key components: a \emph{Distributional Collision Estimator (DCE)} that predicts the distribution of collision possibility, and a modified PPO objective that enhances safety by reducing uncertainty and minimizing collision risk.
Subsequent sections present the task formulation, policy architecture, DCE implementation, and training procedure.

\subsection{Task and Environment Formulation}
\label{sec:task}

We formulate the task as a goal-conditioned non-prehensile manipulation problem in the presence of obstacles, using a mobile manipulator that relies solely on a base-mounted 2D LiDAR.
The sensor provides partial observations, as the manipulated object occludes portions of its field of view—a fundamental constraint that introduces perceptual uncertainty during manipulation.
Fig.~\ref{fig:env} illustrates our simulation environment, detailing the spatial layout for the task, including start, goal, and obstacle regions.
The object and obstacles vary in size, with obstacles appearing at different locations and times during the episode, simulating scenarios such as boxes suddenly falling from shelves or tools left in the workspace.
This variability motivates active perception during manipulation, as reducing uncertainty in occluded regions improves safety.

\subsection{Non-prehensile Manipulation Policy under Uncertainty}
\label{sec:push}
As illustrated in Fig.~\ref{fig:framework}, our manipulation policy $\pi_{\phi}$ maps partial LiDAR observations and the goal pose to robot actions for collision-free object manipulation.
We formulate this as a Partially Observable Markov Decision Process (POMDP) with a policy $\pi_{\phi}(\boldsymbol{a}_t|\boldsymbol{o}_t)$ that maximizes the expected return.

\subsubsection{Sensor Encoding}
We encode the robot's LiDAR observations into a compact spatiotemporal representation that accounts for both current measurements and their reliability over time.
Simply accumulating past LiDAR scans expands spatial coverage but lacks temporal distinction.
This becomes problematic when unforeseen obstacles appear in previously observed but currently occluded regions, treating potentially dangerous areas as safe based on outdated observations.
We address this issue using confidence maps~\cite{ryu2022confidence}, where sensor readings decay exponentially over time—keeping recent measurements reliable while fading outdated ones:
\begin{equation}
M_t(x,y) = \begin{cases}
    1, & \text{if } (x,y) \text{ observed at } t \\
    \alpha \cdot M_{t-1}(x,y), & \text{otherwise}
\end{cases}
\end{equation}
where $M_t$ represents the global confidence map at time step $t$, with values $M_t(x,y) \in [0,1]$ at each pixel ($0.1\mathrm{m}$ resolution) and a decay factor $\alpha \in (0,1)$.
For encoding, we extract a local window $M^{\text{local}}_{t} \in \mathbb{R}^{100 \times 100}$ centered at the robot's base from this global map and encode it into a latent vector $\boldsymbol{z}_t$ using a Variational Autoencoder (VAE) pretrained on random trajectories and frozen during policy learning.

\subsubsection{Policy Architecture}
Our policy $\pi_{\phi}: \Omega \rightarrow \mathcal{A}$ maps the 4-tuple observation $\boldsymbol{o} = (\boldsymbol{o}_p, \boldsymbol{o}_o, \boldsymbol{p}^b_g, \boldsymbol{z})$ to an action.
The observation, expressed in the base frame, consists of:
(1) proprioceptive information $\boldsymbol{o}_p = [\boldsymbol{p}^b_e, \boldsymbol{v}^b_e, \boldsymbol{v}^b_b, \boldsymbol{a}_{t-1}]$ including the end-effector pose and velocity, base velocity, and previous action; 
(2) object information $\boldsymbol{o}_o = [\boldsymbol{p}^b_o, \boldsymbol{v}^b_o]$ with current pose and velocity; 
(3) goal pose $\boldsymbol{p}^b_g$; and 
(4) an encoded confidence map $\boldsymbol{z}$ providing partial environmental information.
The policy outputs velocity commands $\boldsymbol{a} = [\boldsymbol{v}^{b,\text{des}}_b, \boldsymbol{v}^{b,\text{des}}_e]$ for the omnidirectional base and end-effector, which are converted to joint-space targets via differential inverse kinematics and executed through joint impedance control.

\begin{table}[t!]
\caption{Manipulation Policy Reward $\mathcal{R}=\mathcal{R}_{\text{task}}+\mathcal{R}_{\text{reg}}$}
\vspace{-0.1cm}
\label{table:reward}
\renewcommand{\arraystretch}{1.3}
\centering
% \begin{tabularx}{0.9\linewidth}{>{\raggedleft\arraybackslash}p{0.3\linewidth}|>{\arraybackslash}p{0.6\linewidth}}
\begin{tabularx}{0.9\linewidth}{
  >{\raggedleft\arraybackslash}p{0.3\linewidth}
  !{\vrule width \arrayrulewidth}
  >{\arraybackslash}p{0.6\linewidth}
}
    \Xhline{1\arrayrulewidth}
    \textbf{Reward Term} & \textbf{Expression} \\
    \Xhline{3\arrayrulewidth}
    % \hline
    \multicolumn{2}{c}{\textbf{Task Rewards}: \raisebox{0.1ex}{$\mathcal{R}_{\text{task}} = \sum_{k=1}^{4} r_k$}} \\
    \Xhline{1\arrayrulewidth}
    \text{Keypoint} ($r_{1}$) & \raisebox{0.1ex}{$w_{1} \cdot \text{exp}(-0.25 \, ||\boldsymbol{G}^{w}_{g}-\boldsymbol{G}^{w}_{o}||_2)$} \\
    \hline
    \text{Progress} ($r_{2}$) & \raisebox{0.1ex}{$w_{2} \cdot \text{exp}(5.0 \, (\hat{\boldsymbol{v}_{o}^{w}} \cdot \hat{\boldsymbol{p}}^{w}_{o\rightarrow g}-1))$} \\
    \hline
    \text{Velocity} ($r_{3}$) & \raisebox{0.1ex}{$w_{3} \cdot ||\boldsymbol{v}^{w}_{o}||_2 / v_{\text{max}}$} \\
    \hline
    \text{Contact Point} ($r_{4}$) & \raisebox{0.1ex}{$w_{4} \cdot \text{exp}(-0.1 \, ||\boldsymbol{p}^{w}_{r}-\boldsymbol{p}^{w}_{e}||_2)$} \\
\end{tabularx}
% \begin{tabularx}{0.9\linewidth}{>{\raggedleft\arraybackslash}p{0.3\linewidth}|>{\arraybackslash}p{0.6\linewidth}}
\begin{tabularx}{0.9\linewidth}{
  >{\raggedleft\arraybackslash}p{0.3\linewidth}
  !{\vrule width \arrayrulewidth}
  >{\arraybackslash}p{0.6\linewidth}
}
    \Xhline{1\arrayrulewidth}
    \multicolumn{2}{c}{\hspace{-1.23cm}\textbf{Regularization Rewards}: \raisebox{0.1ex}{$\mathcal{R}_{\text{reg}} = \sum_{k=5}^{6} r_k$}} \\
    \Xhline{1\arrayrulewidth}
    \text{Action Rate} ($r_5$) & \raisebox{0.1ex}{$w_{5} \cdot \|\boldsymbol{a}_t - \boldsymbol{a}_{t-1}\|_2$} \\
    \hline
    \text{Smooth Action} ($r_6$) & \raisebox{0.1ex}{$w_{6} \cdot ||\ddot{\boldsymbol{q}}||_{2}$} \\
\end{tabularx}
\begin{tabularx}{0.9\linewidth}{ >{\arraybackslash}X|
                              >{\arraybackslash}X}
  \Xhline{3\arrayrulewidth}
    \multicolumn{2}{c}{\hspace{-.4cm}\textbf{Reward Weights}} \\
  \Xhline{1\arrayrulewidth} 
    \multicolumn{2}{c}{
        \begin{tabularx}{0.9\linewidth}{>{\arraybackslash}X
                                     >{\arraybackslash}X
                                     >{\arraybackslash}X
                                     >{\arraybackslash}X
                                     >{\arraybackslash}X
                                     >{\arraybackslash}X}

         \raisebox{0.1ex}{{\hspace{-2mm}\scriptsize$w_{1}$=$4.0$}} & 
         \raisebox{0.1ex}{{\hspace{-2mm}\scriptsize$w_{2}$=$2.0$}} & 
         \raisebox{0.1ex}{{\hspace{-2mm}\scriptsize$w_{3}$=$1.0$}} & 
         \raisebox{0.1ex}{{\hspace{-2mm}\scriptsize$w_{4}$=$1.0$}} & 
         \raisebox{0.1ex}{{\hspace{-2mm}\scriptsize$w_{5}$=-$0.1$}} & 
         \raisebox{0.1ex}{{\hspace{-2mm}\scriptsize$w_{6}$=-$0.1$}} \\
        \end{tabularx}
    } \\
  \Xhline{1\arrayrulewidth}
\end{tabularx}
\begin{tablenotes}
\setlength{\itemindent}{-0.2cm}
\item[] \textbullet\; $\hat{\cdot}$ denotes a unit vector.
\item[] \textbullet\; $v_\text{max}$ indicates the maximum speed of mobile base.
\item[] \textbullet\; $\boldsymbol{G} \in \mathbb{R}^{24}$ is the keypoint vector proposed in \cite{jeon2023learning}, consisting of the 3D coordinates of the eight object corners.
\item[] \textbullet\; $\boldsymbol{p}_{r}$ is a randomly sampled point on the object surface following \cite{dadiotis2025dynamic}.
\end{tablenotes}
\vspace{-7mm}
\end{table}

\subsubsection{Implementation Details}
We use a ClearPath Ridgeback with a 7-DOF Franka arm and a 2D LiDAR with a $360\degree$ field of view, operating at a maximum base speed $v_{\text{max}}$ of 0.8 m/s.
Training was conducted in Isaac Sim~\cite{mittal2023orbit} with 2048 parallel environments on an NVIDIA RTX 4090 GPU, using Proximal Policy Optimization (PPO)~\cite{schulman2017proximal} to optimize the policy.
The policy operates at 5 Hz, generating velocity commands, while joint impedance control runs at 200 Hz.
The differential IK employs the Damped Least Squares (DLS) method with a damping factor $\lambda = 0.02$.
The confidence map uses an exponential update with $\alpha=0.9$.
Table~\ref{table:reward} summarizes the reward terms used for training.

The keypoint reward minimizes the distance between corresponding object and goal keypoints~\cite{jeon2023learning}, while the contact point reward guides the end-effector toward multiple points sampled on the target surface, encouraging diverse pushing behaviors~\cite{dadiotis2025dynamic}.

\subsection{Distributional Collision Estimator}
\label{sec:dce}
While our manipulation policy can execute pushing actions based on partial observations, it struggles to generate movements that actively reduce occlusion-induced uncertainty.
To address this challenge, we introduce a \emph{Distributional Collision Estimator (DCE)} that models collision possibility as a distribution, enabling optimization of both its expectation \emph{(risk)} and variance \emph{(uncertainty)} during policy learning.

\subsubsection{Distribution of Collision Possibility}

We define collision possibility $C^\pi$ as a random variable representing the discounted accumulation of collision indicators while following policy $\pi_{\phi}$ to capture the urgency of future collisions:
\begin{equation}
C^\pi := \sum_{k=0}^{\infty} \gamma_c^k c_{t+k}, \quad c_t = \mathds{1}_{\text{collision at time } t}\in \{0, 1\}
\label{eq:colvar}
\end{equation}
where $\gamma_c\in(0, 1)$ is a discount factor, set to $0.9$, and $c_t$ indi-\newline cates collision at time $t$.
To obtain the inherent uncertainty arising from partial observability, we approximate the conditional distribution $p(C^\pi|\boldsymbol{o}_t)$ using quantile regression~\cite{koenker2005quantile}.
Our DCE network $f_\varphi: \Omega \rightarrow \mathbb{R}^N$ predicts $N$ quantiles:\\
\begin{minipage}{\columnwidth}
\fontsize{9.5pt}{12.5pt}\selectfont
\begin{equation}
f_\varphi(\boldsymbol{o}_t) = \{z_i(\boldsymbol{o}_t)\}_{i=1}^N \approx \{q_{\tau_i}(C^\pi|\boldsymbol{o}_t)\}_{i=1}^N, \quad \tau_i = \frac{i-0.5}{N}
\label{eq:quantile}
\end{equation}
\vspace{0.01cm}
\end{minipage}
where $q_{\tau_{i}}(C^\pi|\boldsymbol{o}_t)$ denotes the $i$-th quantile of $p(C^\pi|\boldsymbol{o}_t)$.

\subsubsection{Risk and Uncertainty Estimation}

We define \emph{risk} and \emph{uncertainty} as the conditional expectation and variance of $C^\pi$ given the observation $\boldsymbol{o}_t$.
From the DCE's quantile predictions $\{z_i(\boldsymbol{o}_t;\varphi)\}_{i=1}^N$, we estimate the risk and uncertainty:\\
\begin{minipage}{\columnwidth}
\fontsize{9pt}{12.5pt}\selectfont
\begin{align}
\mathcal{R}(\boldsymbol{o}_t) &:= \mathbb{E}_{p(C^\pi|\boldsymbol{o}_t)}[C^\pi] \approx \frac{1} {N}\sum_{i=1}^{N} z_i(\boldsymbol{o}_t;\varphi)\label{eq:risk}\\
\mathcal{U}(\boldsymbol{o}_t) &:= \text{Var}_{p(C^\pi|\boldsymbol{o}_t)}[C^\pi] \approx \frac{1}{N}\sum_{i=1}^{N} \left(z_i(\boldsymbol{o}_t;\varphi) - \mathcal{R}_\varphi(\boldsymbol{o}_t)\right)^2 \label{eq:uncertainty}
\end{align}
\vspace{0.03cm}
\end{minipage} 
We denote these empirical estimates as $\mathcal{R}_\varphi$ and $\mathcal{U}_\varphi$ hereafter.
Risk and uncertainty capture different aspects of collision prediction: high risk indicates likely collisions, whereas high uncertainty reflects unreliable predictions from partial observations.
These complementary metrics guide distinct behaviors of the policy—risk drives collision avoidance while uncertainty promotes active perception to resolve ambiguity.

\subsubsection{Training the Distributional Collision Estimator}

We train the DCE network $f_\varphi$ by minimizing the energy distance between predicted and target quantile distributions~\cite{schneider2024learning}:
\begin{minipage}{\columnwidth}
\fontsize{9.5pt}{12.5pt}\selectfont
\begin{equation}
\mathcal{L}_{\text{DCE}}(\varphi) = 2\mathbb{E}_{i,j}[z_i - \mathcal{T}z_j] - \mathbb{E}_{i,j}[\mathcal{T}z_i - \mathcal{T}z_j] - \mathbb{E}_{i,j}[z_i - z_j]
\label{eq:dce_loss}
\end{equation}
\vspace{0.03cm}
\end{minipage}
where $\mathcal{T}$ denotes the distributional Bellman operator computing target quantiles from observed trajectories.
The energy distance loss promotes distributional consistency: the first term aligns predictions with targets while the remaining terms preserve distributional shape by preventing collapse to point estimates.
During training, DCE parameters $\varphi$ are updated jointly with the policy $\pi_{\phi}$ using $N=50$ quantiles to balance resolution and computational efficiency.

\begin{adjustwidth}{-5cm}{-5cm}

\begin{algorithm}[t] % [H] 같은 플로팅 옵션은 제거합니다.
    \caption{\emph{CURA-PPO}} % 이제 정상적으로 작동합니다.
    \label{alg:cura_ppo}

    % \SetInd 등 algorithm2e 명령어는 algorithm 내용 앞에 와야 합니다.
    \SetInd{0.6em}{0.6em} 
    \For{iteration = 1, 2, \dots}{
        \For{env = 1, 2, \dots, N}{
            Run policy $\pi_{\phi_{\text{old}}}$ for $T$ timesteps\;}
            Compute advantage estimates $\hat{A}_{1\dots T}$\;
            Compute risk cost $\mathcal{C}^{\mathcal{R}}_{1,\dots,T}$ w.r.t $f_{\varphi_{old}}$ \\ \, (Eq.~\ref{eq:cost_risk})\;
            Compute uncertainty cost $\mathcal{C}^{\mathcal{U}}_{1,\dots,T}$ w.r.t $f_{\varphi_{old}}$ \\ \, (Eq.~\ref{eq:cost_uncertainty})\;
        
        Compute CURA policy objective $\mathcal{J}^{CURA}$ (Eq.~\ref{eq:cura_objective}) w.r.t $\pi_\phi$, with $K$ epochs and minibatch size $M$\;
        Update actor, critic and DCE: $\phi_{\text{old}} \leftarrow \phi$, $\varphi_{\text{old}} \leftarrow \varphi$\;
    }
\end{algorithm}
\vspace{2cm}
\end{adjustwidth}
\vspace{-13mm}
\subsection{CURA-PPO: Collision Uncertainty-Risk Aware PPO}
\label{sec:cura}
Building upon the distributional predictions from the DCE, we introduce \emph{CURA-PPO}, augmenting standard PPO with a policy objective that increases the probability of selecting actions that reduce the DCE-estimated risk and uncertainty.
This integration transforms a single-objective task-completion formulation into a multi-objective problem that also targets collision avoidance and uncertainty reduction.

\subsubsection{Cost Formulation from Distributional Statistics}

We formulate two one-step temporal-difference (TD) cost signals from the DCE’s predictions.
The \emph{risk cost} is the TD residual of the expected collision possibility, quantifying how much the current action increases expected future collision beyond the immediate collision.
The \emph{uncertainty cost} is the change in predictive variance, quantifying whether the next observation becomes more or less uncertain than the current one:
\begin{align}
\mathcal{C}^{\mathcal{R}}_t &= \left( c_t + \gamma_c \mathcal{R}_\varphi(\boldsymbol{o}_{t+1}) - \mathcal{R}_\varphi(\boldsymbol{o}_t) \right) \label{eq:cost_risk}\\[0.8ex]
\mathcal{C}^{\mathcal{U}}_t &= \left( \mathcal{U}_\varphi(\boldsymbol{o}_{t+1}) - \mathcal{U}_\varphi(\boldsymbol{o}_t) \right) \label{eq:cost_uncertainty}
\end{align}
where $\boldsymbol{o}_{t+1}$ is the resulting observation after executing action $\boldsymbol{a}_t$ given observation $\boldsymbol{o}_t$.
In the risk cost (Eq.~\ref{eq:cost_risk}), $c_t$ denotes the immediate collision indicator and $\gamma_c$ the discount factor from Eq.~\ref{eq:colvar}.
A negative risk cost means the current action reduces the expected collision possibility.
A negative uncertainty cost (Eq.~\ref{eq:cost_uncertainty}) indicates reduced predictive uncertainty.

\subsubsection{Policy Objective}
We integrate these costs into PPO by augmenting the advantage with penalty terms.
Building on Equation~\ref{eq:ppo_standard}, the \emph{CURA-PPO} objective becomes:
\begin{equation}
\mathcal{J}^{CURA}(\phi) = \mathbb{E}_{\pi_{\phi_{\text{old}}}}\left[\ \rho(\phi) \cdot \left(\hat{A}_t - \lambda_r\bar{\mathcal{C}}^{\mathcal{R}}_t - \lambda_u\bar{\mathcal{C}}^{\mathcal{U}}_t\right)\right]
\label{eq:cura_objective}
\end{equation}
where $\bar{\mathcal{C}}^{\mathcal{R}}_t$ and $\bar{\mathcal{C}}^{\mathcal{U}}_t$ denote normalized costs computed across the batch, weighted by $\lambda_r$ and $\lambda_u$ respectively, with $\lambda_r=0.25$ and $\lambda_u=1.0$.
The probability ratio $\rho(\phi) = \frac{\pi_\phi(\boldsymbol{a}_t|\boldsymbol{o}_t)}{\pi_{\phi_{\text{old}}}(\boldsymbol{a}_t|\boldsymbol{o}_t)}$ and its clipping mechanism are identical to standard PPO.
The subtracted cost terms act as intrinsic penalties, increasing the probability of actions that reduce the expected collision possibility and predictive uncertainty.
Algorithm~\ref{alg:cura_ppo} presents the complete training procedure: rollouts and optimization.

%%%%%%%%%%%%%%%%%%%%%%%%%%%%%%%%%%%%%%%%%%%%%%%%%%%%%%%%%%%%%%%%%%%%%%%%%%%%%%%%

\begin{table*}[t]
\caption{\textbf{Quantitative Manipulation Performance.}
Success rates ($\uparrow$) on 10,000 test cases.}
\vspace{-0.3cm}

\renewcommand{\arraystretch}{1.2}
\begin{center}
% \begin{tabularx}{0.84\textwidth}
% {
%      >{\centering\arraybackslash}p{0.8cm}
%     |>{\arraybackslash}p{3.5cm}
%     |>{\centering\arraybackslash}p{1.2cm}
%      >{\centering\arraybackslash}p{1.2cm}
%      >{\centering\arraybackslash}p{1.2cm}
%     |>{\centering\arraybackslash}p{1.2cm}
%      >{\centering\arraybackslash}p{1.2cm}
%      >{\centering\arraybackslash}p{1.2cm}
% }
\begin{tabularx}{0.84\textwidth}{
  >{\centering\arraybackslash}p{0.8cm}
  % !{\vrule width \arrayrulewidth}
  >{\arraybackslash}p{3.5cm}
  % !{\vrule width \arrayrulewidth}
  >{\centering\arraybackslash}p{1.2cm}
  >{\centering\arraybackslash}p{1.2cm}
  >{\centering\arraybackslash}p{1.2cm}
  !{\vrule width \arrayrulewidth} % ← Uniform / Adversarial 경계
  >{\centering\arraybackslash}p{1.2cm}
  >{\centering\arraybackslash}p{1.2cm}
  >{\centering\arraybackslash}p{1.2cm}
}
\hline
\multirow{2}{*}{Group} & \multicolumn{1}{c|}{\multirow{2}{*}{Method}} & \multicolumn{3}{c@{}}{Uniform} & \multicolumn{3}{c}{Adversarial} \\ \cline{3-8}
& & $0.5^3$ & $0.75^3$ & $1.0^3$ & $0.5^3$ & $0.75^3$ & $1.0^3$ \\ \hline \hline
\multirow{3}{*}{G1} & Push w/o Occlusion & 92.04 & 88.10 & 89.49 & 86.57 & 79.41 & 74.19 \\
\cdashline{2-8}[3pt/4pt]
& Push w/ Occlusion (Baseline)& 74.20 & 66.39 & 42.34 & 65.25 & 48.86 & 22.90 \\
& \emph{CURA-PPO} (Ours) & \textbf{88.73} & \textbf{81.61} & \textbf{80.13} & \textbf{77.12} & \textbf{74.59} & \textbf{72.42} \\ \hline
\multirow{3}{*}{G2} & Baseline w/ Conf. & 80.63 & 68.77 & 63.74 & 69.65 & 63.68 & 54.13 \\ 
& \emph{CURA} w/o Uncertainty  & 84.57 & 69.35 & 64.27 & 70.55 & 64.08 & 57.80 \\
& \emph{CURA} w/o Risk & 86.09 & 79.93 & 78.45 & 78.20 & 70.06 & 67.82 \\ \hline
\multirow{2}{*}{G3} & Push w/ Base & 66.33 & 47.45 & 40.18 & 56.97 & 30.94 & 17.64 \\ 
& \emph{CURA-PPO} w/ Base  & 86.75 & 72.20 & 50.08 & 76.41 & 57.31 & 31.03 \\ \hline
\end{tabularx}

\begin{tablenotes}
\setlength{\itemindent}{1.0cm}
\item[] \textbullet\; $0.5^3$, $0.75^3$, $1.0^3$ denote the size of manipulated cuboid object.
\item[] \textbullet\; Each group (G1, G2, and G3) is distinguished for clarify the explanation.
\end{tablenotes}

\label{table:result}
\vspace{-0.4cm}
\end{center}
\end{table*}

\section{EXPERIMENTAL RESULTS}
We evaluate \emph{CURA-PPO} through extensive experiments assessing its effectiveness in handling object-induced occlusions during non-prehensile manipulation.
Our evaluation includes comparisons with two reference approaches to quantify both the severity of occlusion-induced performance degradation and our method's ability to mitigate this challenge.
We conduct ablation studies analyzing individual component contributions, and qualitative analysis of learned behaviors.
Finally, we examine the advantage of mobile manipulation over base-only pushing for uncertainty reduction.

\subsection{Experimental Setup}
\label{sec:setting}
\begin{figure}[h]
  \centering
  \includegraphics[width=0.95\columnwidth]{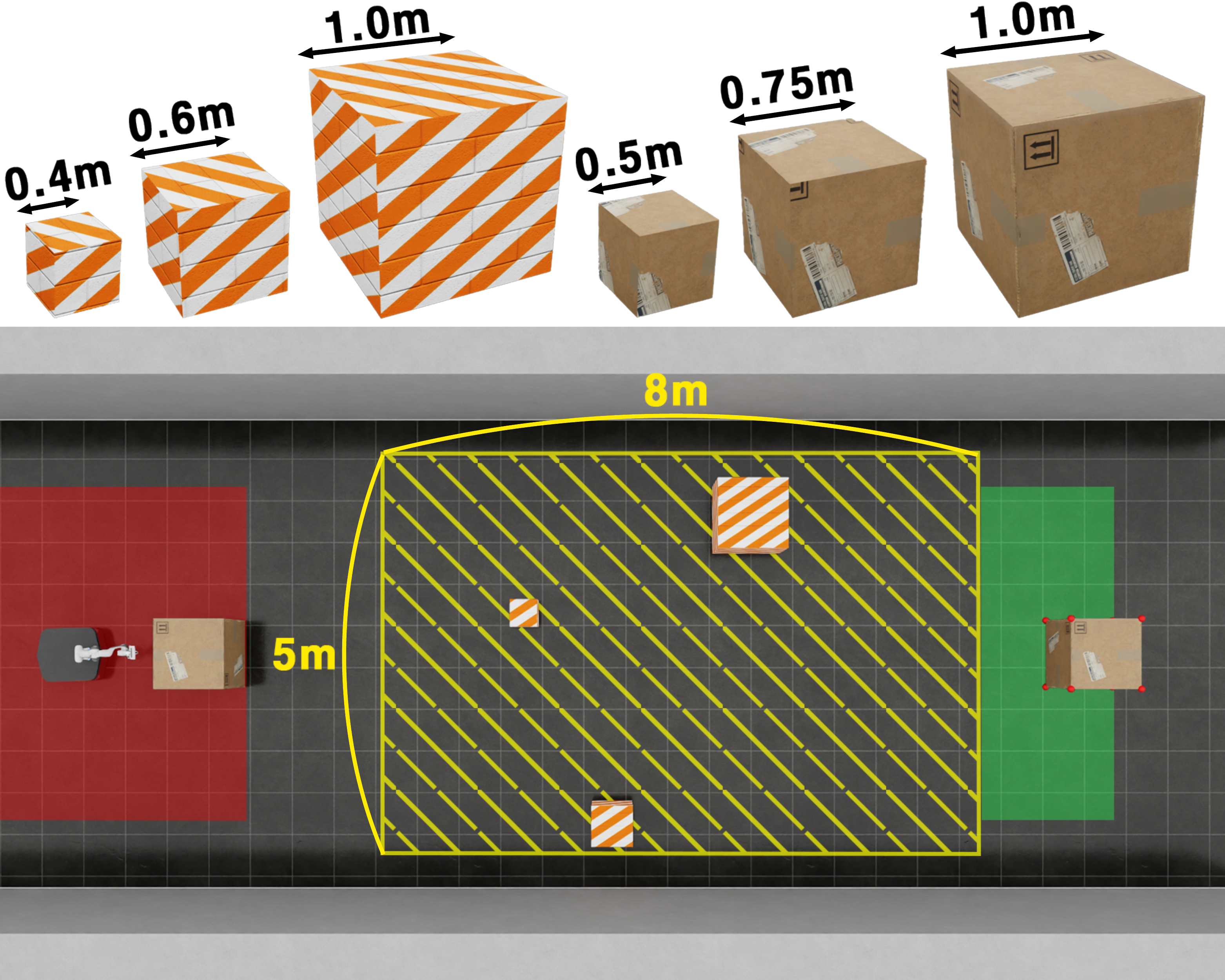}
  \caption{
    \small \textbf{Environment used for training and evaluation.}
    The initial pose for the object is sampled from the \textit{red} region.
    Obstacles are randomly generated within the \textit{yellow} region, and the goal pose is sampled from the \textit{green} region.
    The object and obstacles are cuboids of varying sizes, with 0 to 6 obstacles present in each episode.
    The top-left shows an obstacle, and the top-right shows the object.
    }
  \label{fig:env}
  \vspace{-0.7cm}
\end{figure}

Experiments are conducted in the simulation environment described in Section~\ref{sec:task}, with systematic variations to test performance under various occlusion conditions.
We evaluate three object sizes (0.5$\mathrm{m}$, 0.75$\mathrm{m}$, and 1.0$\mathrm{m}$), where larger objects create more severe occlusions by blocking wider sensor regions.
As shown in Fig.~\ref{fig:env}, obstacles (0.4$\mathrm{m}$, 0.6$\mathrm{m}$, and 1.0$\mathrm{m}$ cuboids) spawn randomly within a region 3-11$\mathrm{m}$ ahead of the initial object position, providing sufficient clearance to avoid immediate collisions.
We test two obstacle scenarios: (1) \emph{Uniform}, where obstacles appear randomly across the workspace, and (2) \emph{Adversarial}, where obstacles are positioned along the direct object-to-goal path, requiring active perception and collision avoidance.
In both scenarios, obstacle positions are randomly spawned at random times to reflect conditions where unexpected obstacles appear during manipulation.
Episodes terminate upon collision or after 50 seconds, with success defined as reaching the goal without collision.
All results are averaged over 10,000 episodes for each combination of object size and obstacle scenario.

\subsection{Evaluation of Manipulation Performance}
\label{sec:performance}

We compare \emph{CURA-PPO} against two reference policies to assess the severity of occlusion-induced performance degradation and validate our method's effectiveness:

\begin{itemize}[leftmargin=0.5cm]
    \item \emph{Push w/o Occlusion}: A policy trained in a setting where all occlusions from the manipulated object are artificially removed by filtering LiDAR ray intersections, relying solely on task rewards and temporal sensor accumulation.
    This serves as an upper bound, representing achievable performance without occlusion constraints.

    \item \emph{Push w/ Occlusion}: A baseline policy trained under realistic occlusion conditions, relying solely on task rewards without confidence maps or explicit uncertainty handling.
\end{itemize}

Table~\ref{table:result} (Group G1) clearly demonstrates the impact of occlusion on manipulation performance and validates our method's effectiveness.
\emph{Push w/o Occlusion} consistently achieves high performance across all object sizes and obstacle scenarios, confirming that successful manipulation is feasible when sensing is unobstructed.
In contrast, \emph{Push w/ Occlusion} exhibits substantial performance degradation, with success rates dropping significantly as object size increases.
This degradation is particularly severe in the \emph{Adversarial} scenario where occlusion severely hinders obstacle detection.

\emph{CURA-PPO} recovers much of the performance lost to occlusion, achieving substantial improvements over the baseline particularly in challenging scenarios.
When occlusion occurs, our method actively reduces uncertainty in occluded regions while minimizing collision risk through distributional estimates from the DCE.
As a result, \emph{CURA-PPO} nearly matches the performance of \emph{Push w/o Occlusion}, demonstrating that explicit risk and uncertainty modeling effectively mitigates manipulation under partial observability.

\subsection{Analysis of Components within CURA-PPO}
\label{sec:ablation}
\vspace{-4mm}

We conduct ablation studies to analyze the individual contributions of three key components—confidence maps, risk cost, and uncertainty cost—with results in Table~\ref{table:result} (G2).

\subsubsection{Evaluation of Confidence Maps}
We evaluate \emph{Baseline w/ Conf.}, a variant that incorporates confidence maps without DCE-based costs.

Confidence maps accumulate past sensor readings while applying temporal decay, enabling it to distinguish recently observed regions from outdated information.
This temporal awareness improves performance over the standard baseline (i.e., \emph{Push w/ Occlusion}), particularly for larger objects where occlusion affects wider regions.
Results indicate that capturing observation reliability provides guidance for decision-making, though the improvement remains limited without explicit uncertainty-driven behaviors.

\subsubsection{Evaluation of Risk and Uncertainty Costs}
We examine the individual effects of risk and uncertainty costs by testing variants that exclude each component while retaining confidence maps.
\emph{CURA w/o Uncertainty} improves upon \emph{Baseline w/ Conf.} but shows limited enhancement compared to \emph{CURA-PPO}, lacking guidance for active perception to resolve occlusions.
\emph{CURA w/o Risk} yields higher success rates, as uncertainty reduction promotes information-seeking behaviors addressing partial observability challenges.

These results reveal a hierarchy in component importance.
Uncertainty cost proves more critical, directly motivating the robot to reduce occlusion through active sensing—a fundamental requirement for safe navigation under partial observability.
Risk cost provides complementary collision avoidance signals that enhance safety margins beyond the task rewards alone.
While task rewards implicitly penalize collisions through episode termination, risk cost offers continuous gradient signals that guide the robot away from potential collisions before critical situations arise.

Combining both costs (i.e., \emph{CURA-PPO}) achieves the best performance, with uncertainty cost driving active perception and risk reducing expected collision possibility.
This complementary effect proves particularly valuable in challenging scenarios with large objects and adversarial obstacles, where both proactive information gathering and collision avoidance are essential for successful manipulation under occlusion.

\begin{figure*}[t]
  \centering
  \includegraphics[width=2.0\columnwidth]{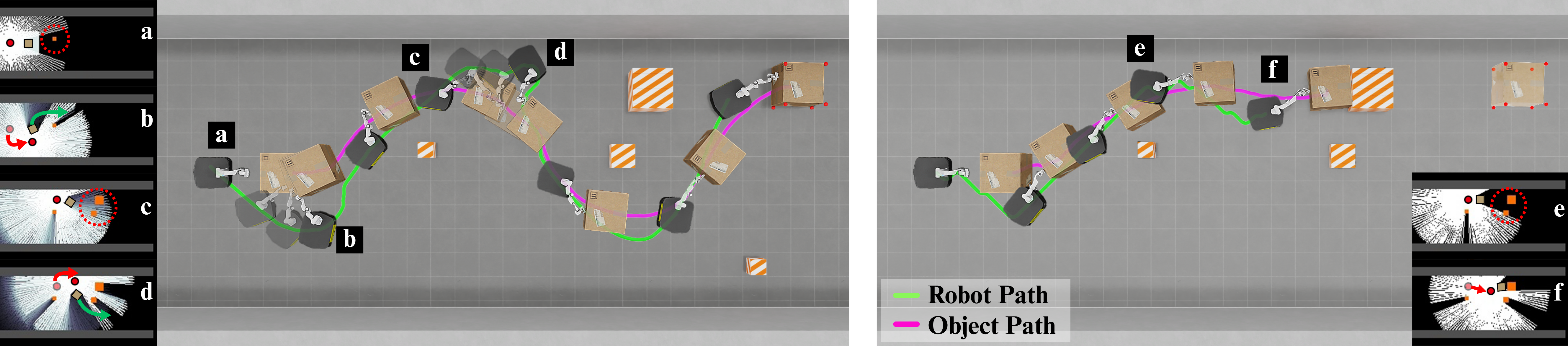}
    \caption{
    \small \textbf{Qualitative Manipulation Behavior under Occlusion}.
    Trajectories shown with robot paths in \emph{green} and object paths in \emph{magenta}.
    (Left) \emph{CURA-PPO} with corresponding confidence maps.
    The robot actively repositions laterally (a→b, c→d) to reveal occluded regions, successfully detecting and avoiding obstacles during pushing the object to the goal.
    (Right) The baseline policy, which lacks both confidence maps and DCE-based costs, fails to explore occluded regions, resulting in collision with an undetected obstacle that was hidden behind the manipulated object (e→f).
    Without these uncertainty-aware mechanisms, the robot maintains a direct path that leads to task failure.
    }    
  \label{fig:qualitative}
  \vspace{-0.6cm}
\end{figure*}

\subsection{Qualitative Analysis of Manipulation Behavior}
\label{sec:qualitative}

\begin{figure}[t]
  \centering
  \includegraphics[width=1.0\columnwidth]{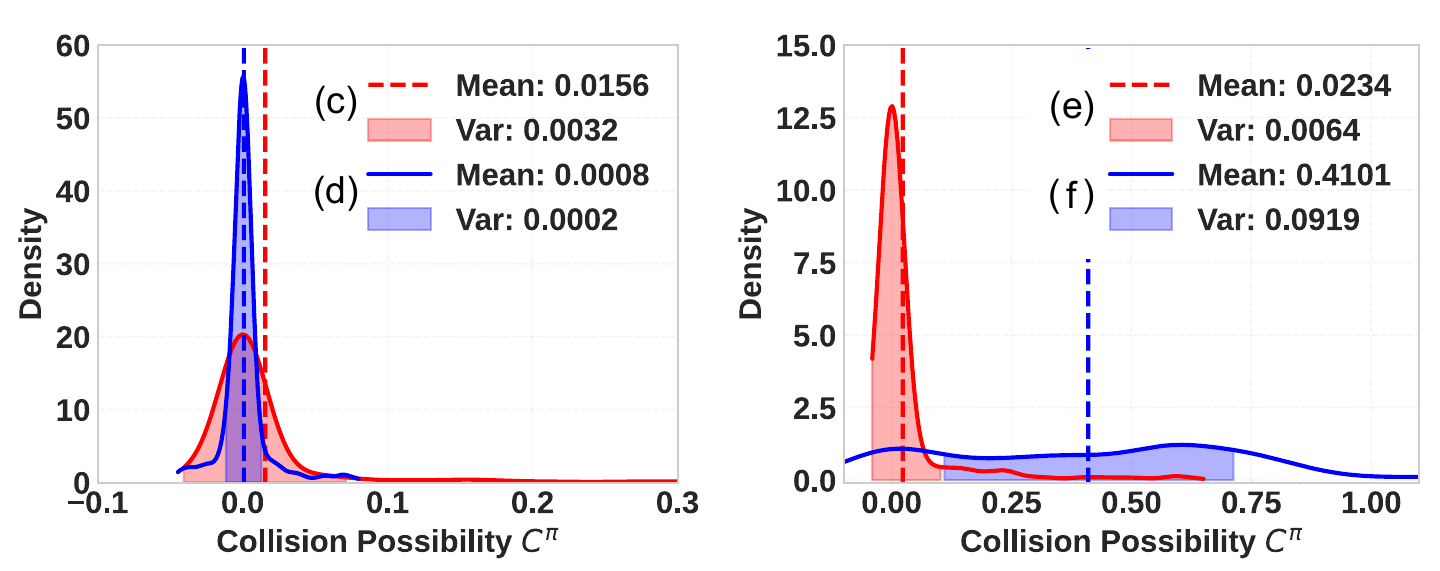}
    \caption{
        \small \textbf{Visualization of DCE predictions} at time steps from Fig.~\ref{fig:qualitative}. (Left) \emph{CURA-PPO} reduces variance from c[\textit{red}] to d[\textit{blue}] through active perception.
        (Right) Baseline maintains high variance from e[\textit{red}] to f[\textit{blue}], leading to collision.
    }    
  \label{fig:dce}
  \vspace{-0.7cm}
\end{figure}

We analyze manipulation trajectories and corresponding DCE predictions (Figs.~\ref{fig:qualitative} and \ref{fig:dce}) to examine how our method develops active perception strategies under occlusion, comparing \emph{CURA-PPO} against the baseline policy (i.e., \emph{Push w/ Occlusion}) to reveal distinct manipulation behavior.

Fig.~\ref{fig:qualitative} illustrates the contrasting manipulation strategies between \emph{CURA-PPO} and the baseline.
In sequences (a→b) and (c→d), \emph{CURA-PPO} responds to low-confidence regions in its perception map by executing lateral movements deviating from the direct path, revealing undetected obstacles while maintaining object control.
These uncertainty-reduction maneuvers, though seemingly inefficient for immediate progress, enable obstacle avoidance during manipulation.
The baseline policy (e→f), lacking uncertainty signals, maintains direct pushing toward the goal and collides with the unexpected obstacle hidden behind the manipulated object.
This behavioral difference highlights how uncertainty-aware decision-making influences manipulation strategy.

Complementing Fig.~\ref{fig:qualitative}, the DCE predictions in Fig.~\ref{fig:dce} provide quantitative evidence for uncertainty-aware active-perception.
At (c), the estimated distribution shows higher variance due to limited information.
Our approach reduces variance at (d) via uncertainty-reducing actions, indicating active perception lowers DCE uncertainty in collision possibility.
In contrast, the baseline faces a similar situation at (e) but proceeds without DCE guidance.
This single-objective strategy keeps variance high through (f), leading to a collision.
These observations suggest DCE predictions provide reliable guidance for safe manipulation under occlusion.

\begin{figure}[t]
  \centering
  \includegraphics[width=1.0\columnwidth]{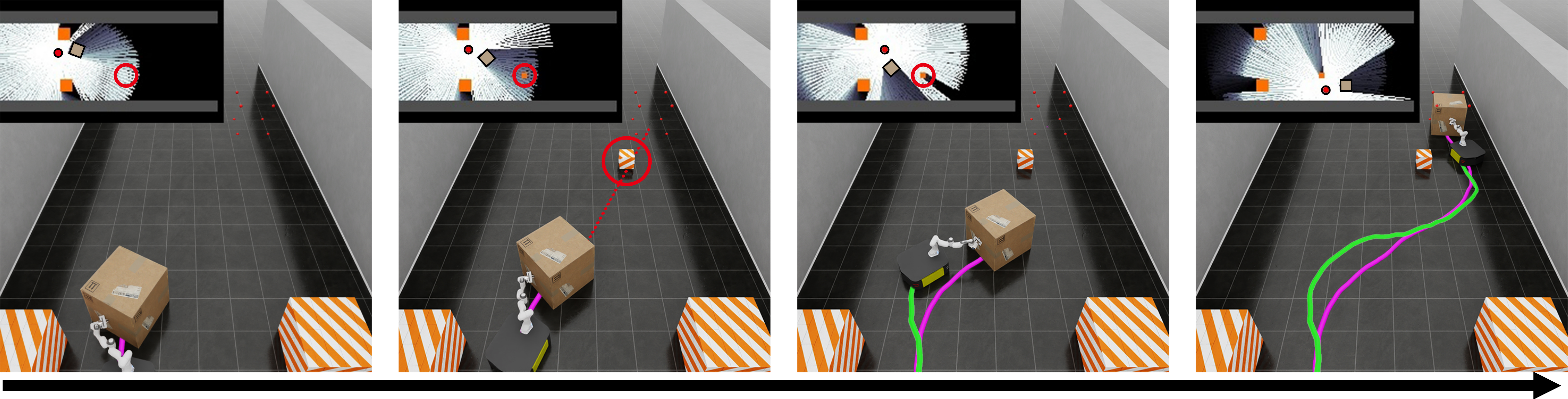}
    \caption{
    \small \textbf{Manipulation Behavior in \emph{Adversarial} Scenario}.
    Sequential snapshots (left to right) demonstrate \emph{CURA-PPO}'s response to adversarial obstacles that suddenly appear in previously clear regions. \emph{Green} and \emph{magenta} paths show robot and object trajectories.
    }
  \label{fig:adversarial}
  \vspace{-0.5cm}
\end{figure}

The active perception capabilities of \emph{CURA-PPO} prove particularly important in the \emph{Adversarial} scenario, where obstacles appear along the direct pushing path.
Fig.~\ref{fig:adversarial} shows how obstacles emerging in previously observed regions create a critical challenge—they become hidden behind the manipulated object during pushing.
\emph{CURA-PPO} successfully addresses this challenge through its uncertainty-aware strategy: confidence maps signal when spatial information becomes unreliable, triggering lateral movements that reveal the occluded obstacle. This proactive detection enables the robot to adjust its pushing direction in real-time, guiding the object safely around the obstacle to the goal.

\subsection{Role of Kinematic Redundancy in Manipulation}
\label{sec:mobile}
We conducted an additional ablation study using a base-only platform (without a manipulator arm) to evaluate how kinematic redundancy affects manipulation performance under occlusion.
Table~\ref{table:result} (G3) presents two configurations trained under occlusion: \emph{Push w/ Base}, a base-only agent with task rewards, and \emph{CURA-PPO w/ Base}, which applies our uncertainty-aware framework to the same platform.

\begin{figure}[t]
  \centering
  \includegraphics[width=1.0\columnwidth]{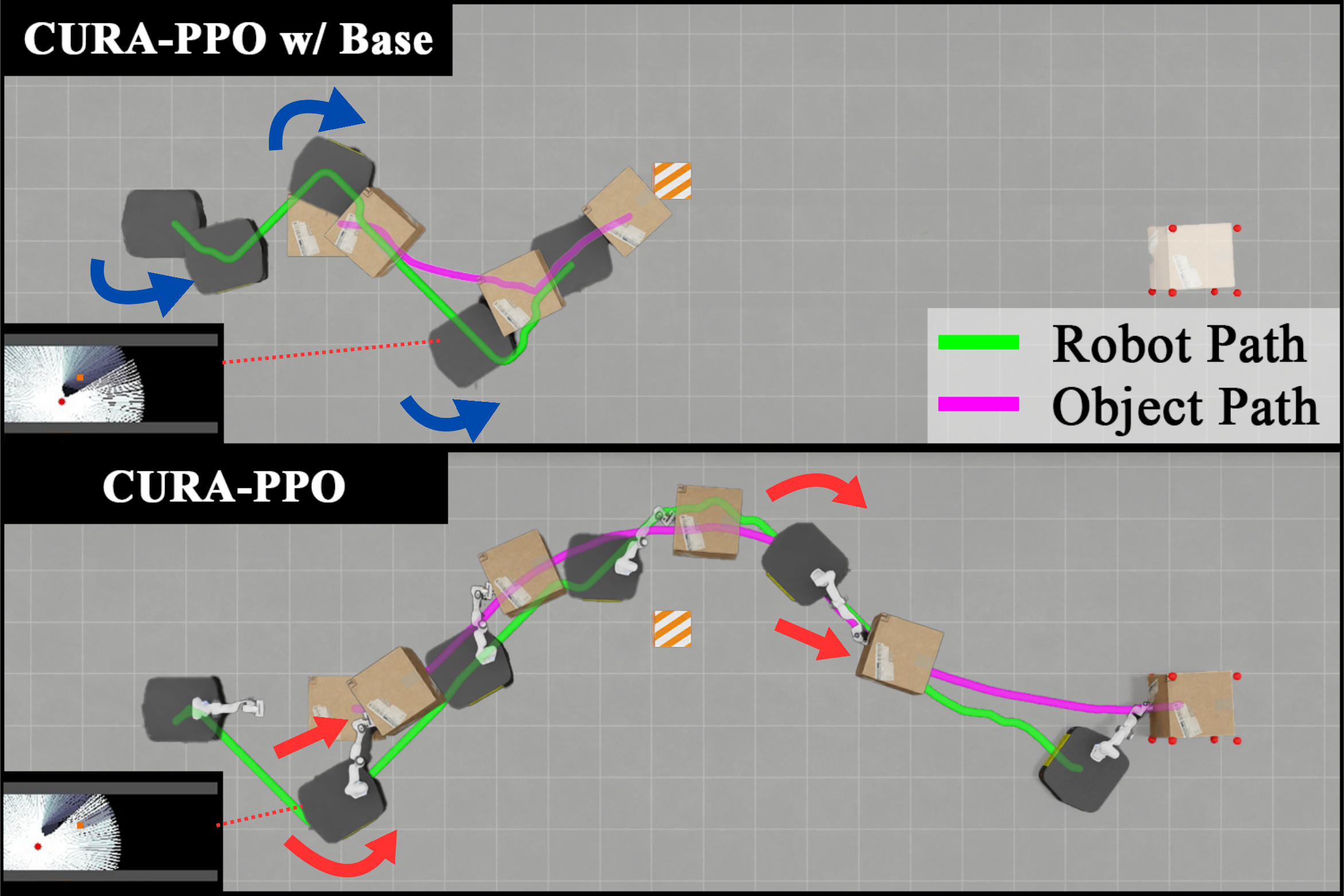}
    \caption{
    \small \textbf{Behavioral Comparison of Base-Only and Mobile Manipulator Systems}.
    (Top) Base-only platform fails due to restricted viewing angle, resulting in insufficient visual coverage while pushing the object to the target.
    (Bottom) Mobile manipulator successfully leverages kinematic redundancy to maintain object contact while adjusting viewing angles for effective perception.
    }
  \label{fig:basepush}
  \vspace{-0.5cm}
\end{figure}

The results show that base-only manipulation consistently underperforms compared to configurations with a manipulator arm.
In the absence of kinematic redundancy, the robot has limited ability to reduce object-induced occlusion while maintaining contact and therefore often advances while the object’s forward path in or near unobserved regions.
Although \emph{CURA-PPO w/ Base} improves over the base-only baseline, these benefits decline as object size increases and occlusion becomes more severe.
As shown in Fig.~\ref{fig:basepush}, this example illustrates a failure of the base-only approach.
By contrast, leveraging kinematic redundancy enables the mobile manipulator to maintain contact while improving visibility, which leads to successful obstacle detection and avoidance.

\vspace{3mm}
For a more intuitive understanding of the experimental
results, please refer to the supplementary video.
\vspace{40mm}

%%%%%%%%%%%%%%%%%%%%%%%%%%%%%%%%%%%%%%%%%%%%%%%%%%%%%%%%%%%%%%%%%%%%%%%%%%%%%%%%

\section{CONCLUSION}
We presented \emph{CURA-PPO}, a novel reinforcement learning framework that addresses the challenge of object-induced occlusion in non-prehensile manipulation using only onboard sensing.
By modeling collision possibility as a distribution through the \emph{Distributional Collision Estimator (DCE)}, our approach extracts both \emph{risk} and \emph{uncertainty} signals to augment standard policy optimization.
These complementary costs encourage distinct yet synergistic behaviors: \emph{risk} drives collision avoidance while \emph{uncertainty} promotes active perception to resolve occlusions during manipulation.
Experiments across varying object sizes and obstacle configurations demonstrate that \emph{CURA-PPO} achieves robust performance in both \emph{Uniform} and \emph{Adversarial} scenarios.
Qualitative analysis reveals that the learned policy strategically maneuvers to gather environmental information while maintaining object control, mitigating the occlusion challenge. 
Future work will focus on deploying this approach on real robot systems and extending the framework to incorporate additional sensor modalities such as RGB-D cameras for enhanced environmental perception in complex real-world scenarios.
\vspace{5mm}
%%%%%%%%%%%%%%%%%%%%%%%%%%%%%%%%%%%%%%%%%%%%%%%%%%%%%%%%%%%%%%%%%%%%%%%%%%%%%%%%
% \input{contents/07_appendix}
%%%%%%%%%%%%%%%%%%%%%%%%%%%%%%%%%%%%%%%%%%%%%%%%%%%%%%%%%%%%%%%%%%%%%%%%%%%%%%%%
% \input{contents/08_acknowledgement}
%%%%%%%%%%%%%%%%%%%%%%%%%%%%%%%%%%%%%%%%%%%%%%%%%%%%%%%%%%%%%%%%%%%%%%%%%%%%%%%%

\small
\bibliographystyle{ieee}
\bibliography{./ref}
%%%%%%%%%%%%%%%%%%%%%%%%%%%%%%%%%%%%%%%%%%%%%%%%%%%%%%%%%%%%%%%%%%%%%%%%%%%%%%%%
\end{document}